%% file: root.tex
\documentclass[letterpaper, 10 pt, conference]{ieeeconf}  

\let\labelindent\relax

\IEEEoverridecommandlockouts                              

\input{config}

\title{\LARGE \bf
Closed-Loop Open-Vocabulary Mobile Manipulation with GPT-4V
}

\author{Peiyuan Zhi$^{1,*}$, Zhiyuan Zhang$^{1,2,*}$, Yu Zhao$^{1 }$, Muzhi Han$^{3}$, Zeyu Zhang$^{1}$, Zhitian Li$^{1}$, \\ Ziyuan Jiao$^{1}$, Baoxiong Jia$^{1,\dagger}$, Siyuan Huang $^{1,\dagger}$ \\
\textbf{\url{https://come-robot.github.io}}
\thanks{$^{*}$ Peiyuan Zhi and Zhiyuan Zhang contributed equally to this work.}
\thanks{
$^\dagger$ Corresponding authors. Emails: 
{\fontsize{7pt}{8.4pt}\selectfont \tt{\{jiabaoxiong,syhuang\}@bigai.ai}}
}%
\thanks{%
    $^{1}$ State Key Laboratory of General Artificial Intelligence, Beijing Institute for General Artificial Intelligence (BIGAI).
    $^{2}$ Department of Automation, Tsinghua University.
    $^{3}$ University of California, Los Angeles.
}
}

\begin{document}

\maketitle
\thispagestyle{empty}
\pagestyle{empty}

\input{sec/0_abstract}

\input{sec/1_intro}

\input{sec/2_related_work}

\input{sec/3_method}
\input{sec/4_experiment}

\input{sec/5_conclusion}

\bibliographystyle{IEEEtranS} 
\section{ACKNOWLEDGMENTS}
We thank Unitree Robotics for their help with the B2 and Z1 robots.
\bibliography{reference}

\end{document}

%% file: config.tex
\usepackage{bm}
\usepackage{mdframed}
\usepackage{acronym}
\usepackage{xspace}
\usepackage{graphicx}
\usepackage{caption}
\usepackage{amsmath}
\usepackage{xcolor}
\usepackage{subcaption}
\usepackage{booktabs}
\usepackage{comment}
\usepackage{tabularx}
\usepackage{multirow}
\usepackage{url}
\usepackage{enumitem}
\usepackage[pagebackref,breaklinks,colorlinks=true,linkcolor=cvprblue,citecolor=cvprblue,bookmarks=false]{hyperref}

\usepackage[capitalise,nameinlink]{cleveref}

\acrodef{llm}[LLM]{Large Language Model}
\acrodef{vlm}[VLM]{Vision Language Model}
\acrodef{ovmm}[OVMM]{Open-Vocabulary Mobile Manipulation}
\acrodef{cot}[CoT]{Chain-of-Thought}
\acrodef{slam}[SLAM]{simultaneous localization and mapping}
\acrodef{tamp}[TAMP]{task and motion planning}
\acrodef{urdf}[URDF]{unified robot description format}
\acrodef{pt}[\emph{pt}]{parse tree}
\acrodef{iou}[IoU]{intersection over union}
\acrodef{map}[mAP]{mean average precision}
\acrodef{cap}[CaP]{Code as Policies}

\let\oldnl\nl
\newcommand{\nonl}{\renewcommand{\nl}{\let\nl\oldnl}}
\newcolumntype{x}{>{\columncolor{MistyRose}}c}
\newcolumntype{y}{>{\columncolor{LightCyan1}}c}
\newcommand{\robot}{COME-robot\xspace}
\newcommand{\sota}{state-of-the-art\xspace}

\definecolor{gblue}{HTML}{4285F4}
\definecolor{gred}{HTML}{DB4437}
\definecolor{ggreen}{HTML}{0F9D58}

\newcommand{\etc}{\emph{etc.}\xspace}
\newcommand{\eg}{\emph{e.g.}\xspace}
\newcommand{\ie}{\emph{i.e.}\xspace}

%% file: sec/0_abstract.tex
\begin{abstract}

Autonomous robot navigation and manipulation in open environments require reasoning and replanning with closed-loop feedback. In this work, we present \robot, the first closed-loop robotic system utilizing the GPT-4V vision-language foundation model for open-ended reasoning and adaptive planning in real-world scenarios. 
\robot incorporates two key innovative modules: (i) a multi-level open-vocabulary perception and situated reasoning module that enables effective exploration of the 3D environment and target object identification using commonsense knowledge and situated information, and (ii) an iterative closed-loop feedback and restoration mechanism that verifies task feasibility, monitors execution success, and traces failure causes across different modules for robust failure recovery.
Through comprehensive experiments involving 8 challenging real-world mobile and tabletop manipulation tasks, \robot demonstrates a significant improvement in task success rate ($\sim$35\%) compared to state-of-the-art methods. We further conduct comprehensive analyses to elucidate how \robot's design facilitates failure recovery, free-form instruction following, and long-horizon task planning.




\end{abstract}

%% file: sec/1_intro.tex

\definecolor{lightblue}{RGB}{0,176,240}
\definecolor{lightred}{RGB}{255,0,0}
\definecolor{deepgreen}{RGB}{0,176,80}
\definecolor{cvprblue}{RGB}{0,113,188}

\section{Introduction}\label{sec:intro}

Developing autonomous robots capable of understanding, navigating, and manipulating objects in unknown real-world environments has been a central focus of robotics research. Recent advances in large foundation models~\cite{brown2020language,bommasani2021opportunities, radford2021clip, gpt4v}, particularly in aligning machine perception with natural language in an open-vocabulary, zero-shot manner, have sparked even more ambitious goals. One such goal is the task of \ac{ovmm}, which evaluates a robot's ability to follow natural language instructions to perform tasks such as navigation and mobile manipulation. However, such expectations bring significant challenges: 
\begin{itemize}[leftmargin=*,nolistsep,noitemsep]
    \item The demand to \textit{perceive and understand unstructured, complex 3D environments} and to align them with natural language for task-oriented planning and reasoning.
    \item The difficulty of \textit{closed-loop replanning} for failure recovery using environmental feedback to successfully \textit{execute long-horizon tasks} in mobile manipulation.
\end{itemize}
Addressing the first challenge requires robotic systems to effectively structure and ground their understanding of the 3D environment for reasoning, while the second challenge demands robust systems that can effectively diagnose failures and efficiently recover from different types of errors for reliable planning and execution.

\begin{figure}[t]
  \centering
  \includegraphics[width=0.95\linewidth]{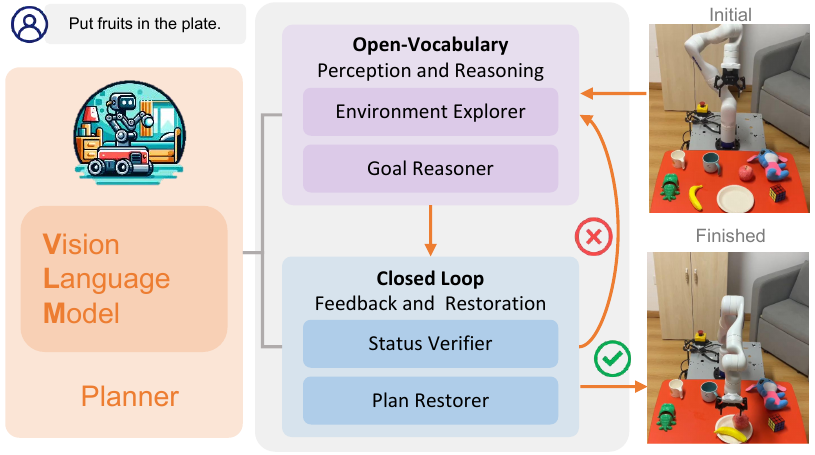}
  \caption{\textbf{Overview of \robot's system.} Given a task instruction, \robot employs GPT-4V for reasoning and generates a code-based plan. Through feedback obtained from the robot's execution and interaction with the environment, it performs closed-loop replanning and iteratively updates the subsequent plan or recovers from failures, ultimately accomplishing the given task.
  }
  \label{fig:teaser}
\end{figure}

Traditional closed-loop replanning systems~\cite{garrett2020online,curtis2022long} rely on predefined symbolic logic for reasoning and replanning, grounding input observations (\eg, images or point clouds) to pre-defined logical symbols and predicates for reasoning. This design significantly restricts the adaptability of robotic systems in open-ended real-world scenarios, where robots must plan in unseen situations. Large Language Models (\acs{llm}s) provide promising approaches~\cite{huang2022inner,yao2022react,liu2023reflect, hu2023look, skreta2024replan} to enhancing robots' ability to elucidate failures, propose corrections, and devise new plans based on textual environmental feedback. With code generation capabilities~\cite{liang2023code,singh2023progprompt}, these models can interface task planning with actual execution by leveraging API functions for robots' primitive actions. Vision-Language Models (\acs{vlm}s) further extend these abilities by enabling open-vocabulary understanding of raw visual observations from robot sensors~\cite{radford2021clip,du2022survey}. However, the applicability of this stream of approach remains largely confined to tabletop settings, where the challenge of partial observability during environment exploration and understanding are mostly unaddressed. This limitation reduces the effectiveness of current closed-loop replanning mechanisms in open-ended, real-world environments, particularly for long-horizon mobile manipulation tasks.

To address the challenges of \ac{ovmm} and the limitations of existing methods, we propose \robot (\textbf{C}losed-loop \textbf{O}pen-vocabulary \textbf{M}obil\textbf{E} Manipulation), the first closed-loop robotic system that integrates GPT-4V~\cite{gpt4v} 
with robust robotic primitive actions for real-world \ac{ovmm}. The model framework of \robot is illustrated in~\cref{fig:teaser}. Building upon prior practices that use \ac{vlm}s for planning, we introduce two key innovations in \robot to address the limitations of existing methods: (i) \textbf{a multi-level open-vocabulary perception and situated reasoning module} for effectively exploring the environment, organizing environment information, translating human instructions to different levels of goals within the scene, and identifying the corresponding target objects, and (ii) \textbf{an iterative closed-loop feedback and restoration mechanism} that identifies failures in task planning and robot execution, considers dependencies between modules (\eg, navigation and local manipulation), and traces the causes of failures across different modules for failure recovery. To evaluate our design, we conduct comprehensive real-robot experiments in a real-world bedroom involving 8 challenging \ac{ovmm} tasks, demonstrating that \robot significantly outperforms existing \acs{llm}-based robotic systems in all tasks. We further provide detailed analyses to identify common failure modes of the system and offer a holistic view on \robot's ability to recover from failures. 

In summary, our contributions are:
\begin{enumerate}[leftmargin=*,nolistsep,noitemsep]
    \item We present \robot, the first closed-loop robotic system capable of open-ended reasoning, navigation, manipulation, and failure recovery for long-horizon \ac{ovmm} in complex real-world environments.
    \item We propose two key modules (i) the open-vocabulary perception and reasoning module, which enables effective interpretation of human instructions as goals with varying granularity in complex real-world environments, and (ii) the closed-loop feedback and recovery module, which autonomously manages planning and execution failures, performing multi-level backtracking to ensure efficient and effective failure recovery.
    \item We conduct comprehensive real-robot experiments, demonstrating that our design significantly outperforms existing \sota methods. Additionally, we offer extensive discussions on emergent robot behaviors, such as failure recovery, exhibited by \robot. 
\end{enumerate}

%% file: sec/2_related_work.tex
\section{Related Work}

\textbf{Open vocabulary mobile manipulation} presents significant challenges in robotics research, requiring the integration of mobile navigation, manipulation, as well as the ability to understand and complete diverse long-horizon tasks in large scenes. This introduces significant complexities~\cite{wolfe2010combined,jiao2021efficieint,yokoyama2023asc}. While many approaches~\cite{brohan2022rt,brohan2023rt,kim2024openvla} use end-to-end models to perform robotic manipulation tasks, these systems are typically limited to handling short-sequence tasks in relatively simple table-top settings due to the complexity of action spaces and environment understanding. Consequently,  current methods rely on high-level task planners and \ac{llm}s to generate plans, which are then executed by robots' available primitive skills~\cite{yenamandra2023homerobot,rana2023sayplan,qiu2024open}. However, as shown by OK-Robot~\cite{liu2024ok} on the HomeRobot~\cite{yenamandra2023homerobot} \ac{ovmm} benchmark, a naive combination of \ac{vlm}-based high-level planners with low-level executors often leads to sequential errors across modules. This underscores the need for effective recovery mechanisms, especially in long-horizon mobile manipulation tasks.

\textbf{Closed-loop robot systems incoporating LLMs and VLMs} have emerged as a promising direction to addressing challenges in open-ended reasoning and planning in robotics~\cite{zeng2023large,liang2023code,singh2023progprompt,wang2023voyager, wake2023gpt, hu2023look, rana2023sayplan}. These systems typically integrate perceptual inputs with \ac{llm}s and \ac{vlm}s for common-sense reasoning and instruction following, and use \ac{llm}-generated code to drive robot execution. Existing approaches often assume flawless low-level relying on environmental feedback primarily for high-level task feasibility verification and refinement~\cite{rana2023sayplan}, overlooking execution failures. Recent approaches mitigate this gap by incorporating success detection and human interaction for failure analysis and replanning at both the plan and execution levels~\cite{yao2022react, guo2023doremi, wang2023describe, wang2023voyager, ding2023integrating, liu2023reflect, hu2023look, skreta2024replan, mei2024replanvlm, han2024interpret}. However, these methods typically use a simple re-attempt strategy for solving execution failures, neglecting the potential interdependencies between modules (\eg, navigation to a narrow side of the table limits access to objects at its center). Therefore, in \robot, we address this issue by focusing on interconnection between different modules and tracing the causes of failures across them for more effective failure recovery.

%% file: sec/3_method.tex
\section{Method}

\begin{figure*}[t]
    \centering
    \includegraphics[width=0.95\linewidth]{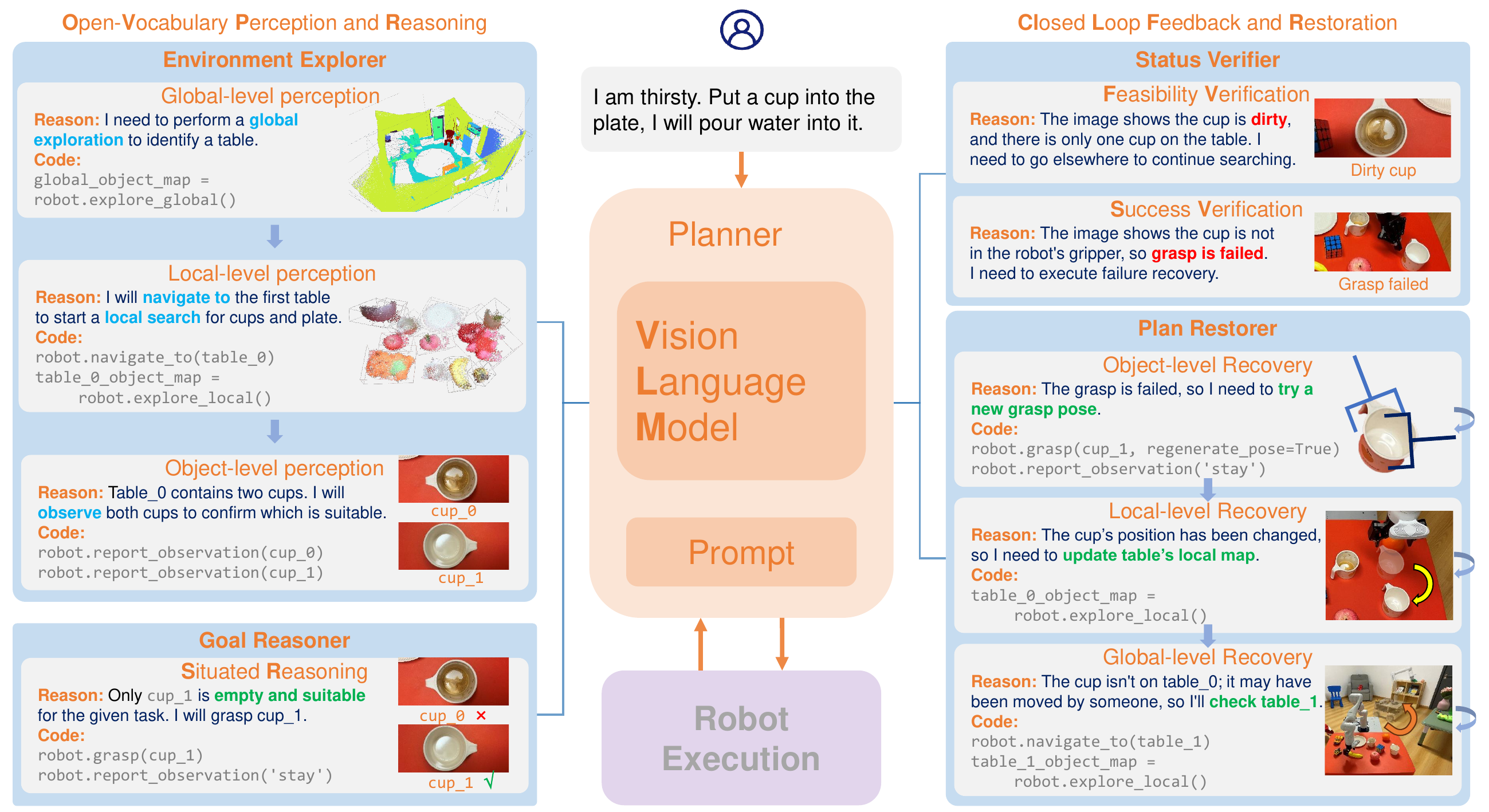}
    \caption{\robot's planner has two key designs: \textbf{Open-Vocabulary Perception and Reasoning} and \textbf{Closed Loop Feedback and restoration}. The former helps the robot ground open-ended instructions in real environment, and the latter 
    guarantees task's completion. Actions to be executed as reasoned by GPT-4V are highlighted in \textcolor{lightblue}{blue}, identified failures are highlighted in \textcolor{lightred}{red}, and analysis after observation or verification are highlighted in \textcolor{deepgreen}{green}.}
    \label{fig:overview}
\end{figure*}

In this section, we detail \robot's approach to closed-loop, open-vocabulary mobile manipulation. Utilizing GPT-4V with robust robotic primitive actions as a foundation, we elucidate on two key modules of \robot: (i) the multi-level open-vocabulary perception and situated reasoning module (\cref{subsec:perception}), and (ii) the closed-loop feedback and restoration mechanism (\cref{subsec:feedback}). We provide an illustrative overview of our framework in~\cref{fig:overview}. Additionally, we provide an overview of the system prompt and implementation in \cref{subsec:implementation}. Please refer to the \textit{supplmentary} on our website for full prompts and additional implementation details.

\subsection{Open-Vocabulary Perception and Situated Reasoning}\label{subsec:perception}

\textbf{Environment Explorer{\ }} To address the innate complexity of large scenes in the mobile manipulation setting, we propose to organize the exploration and understanding of environments at the three different levels:
\begin{itemize}
     \item \textit{Global-level perception}: When entering an unknown environment, \robot first activates global-level perception. This stage of perception uses the robot's mobile base with a frontier-based exploration strategy to scan the environment and build a global object map of large furniture items. For each object, \robot records its category, location, merged point cloud, visual features, and language description. 
     \item \textit{Local-level perception}: After building the global object map, \robot reasons about potential target regions following human task instructions. We leverage GPT-4V's commonsense reasoning ability for locating the target object (\eg, cup should be on table). Once the target region is identified, \robot navigates to that area (\eg, table) and starts constructing a local object placement map of small objects on large receptacle objects for future interactions. Notably, we employ SAM-2~\cite{ravi2024sam} for tracking object status and re-execute local-level perception during task execution if object positions are altered.
     \item \textit{Object-level perception}: In this stage, \robot captures close-up images of potential target objects in the area and utilizes GPT-4V to extract fine-grained object attributes for future reasoning and goal identification.
     
\end{itemize}

{\textbf{Goal Reasoner}{\ }} 
After exploring the scene following the natural language task instruction, we identify the goal object by instructing GPT-4V to reason out the target objects. This process involves commonsense reasoning on object functionality and availability for task execution. For example, as shown in~\cref{fig:overview}, \robot observes two cups on the table and infers that the second cup, being empty, is more suitable for pouring water and therefore passes out a function call for grasping the second cup.

\subsection{Closed Loop Feedback and Restoration}\label{subsec:feedback}

{\textbf{Status Verifier}{\ }}
The status verifier comprises two components: feasibility verification and success verification. Feasibility verification assesses whether the task can proceed or if a rollback is required, using perception feedback from the multi-level environment explorer and the \ac{vlm} dialogue context. For example, when no empty cup is found on the current table, \robot will roll back to search for another table that might have one. Success verification evaluates whether the robot's execution is successful using multi-modal feedback, including the object images captured by the wrist camera before and after execution, as well as feedback from the applied action function calls. It also analyzes the cause of failure to determine the appropriate recovery level (\eg, plan or execution).

{\textbf{Plan Restorer}{\ }}
To recover from unfeasible situations that require a rollback, as well as various execution and perception failures, we propose a hierarchical recovery mechanism to ensure task progress. This mechanism aligns with the hierarchical structure of the environment explorer.

\begin{itemize}
      \item \textit{Object-level recovery}: When \robot encounters a manipulation failure, it first initiates object-level recovery by re-attempting the manipulation action.  For instance, if a grasp failure is detected, the robot will adjust to a different grasping pose and try again. If the object has moved during failed attempts, the status verifier compares images captured before each attempt to reason that the object has shifted. In such cases, \robot rolls back to higher level recovery.
     \item \textit{Local-level recovery}: When object-level recovery is insufficient for successful execution, such as when objects move during the task, the local-level recovery mechanism relocates and reclassifies the target object by updating the local object map and executing a new action. For instance, as shown in \cref{fig:overview}, \robot successfully relocates and grasps the cup. 
     \item \textit{Global-level recovery}: When the plan becomes unfeasible at the current state, or if the robot is unable to interact with the target object from its current position, as verified by the status verifier, \robot initiates global-level recovery. In this phase, \robot navigates to a new target position, seeking alternative furniture or better access to the target object, allowing it to reattempt the task under improved conditions.
\end{itemize}

\begin{figure}[t]
  \centering
  \includegraphics[width=0.86\linewidth]{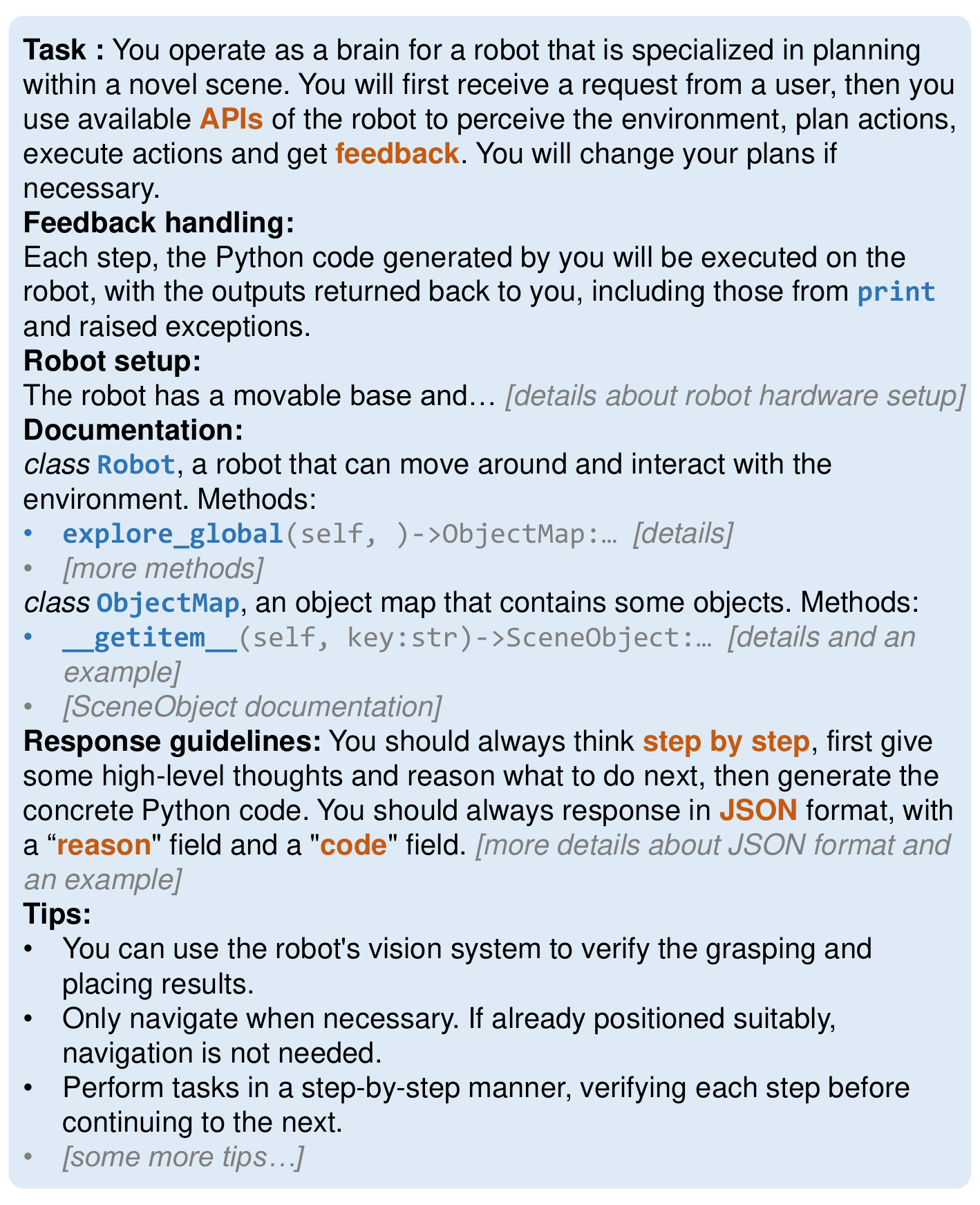}
  \caption{\textbf{A snapshot of \robot's system prompt.}}
  \label{fig:prompt}
\end{figure}

\subsection{Implementation Details}\label{subsec:implementation}
\paragraph*{Prompt} 
We prompt GPT-4V for planning and failure recovery. 
As shown in \cref{fig:prompt}, the system prompt consists of task description, feedback handling, details about robot setup and skills library, response guidelines and useful
tips.
The prompt and user query are fed into GPT-4V, which generates Python code for robot to execute.
\paragraph*{Primitive Actions as APIs}

We identify a library of primitive actions necessary for \robot to solve the \ac{ovmm} task in the closed loop. We categorize them into two types of APIs and detail our design choice in the following. 


\paragraph*{Perception APIs} We design perception APIs \texttt{explore\_global} and \texttt{explore\_local} for constructing the global and local object map, respectively. To provide GPT-4V with visual feedback, we introduce an additional API \texttt{report\_observation} that captures a 2D image using the robot's wrist camera.

\paragraph*{Execution APIs} 
For navigation, we define an API \texttt{navigate} that drives the robot to approach an object that appears in the global object map. In terms of manipulation, we design a \texttt{grasp} API that commands the robot to grasp an object based on observed object point cloud, and a \texttt{place} API that allows the robot to place the object in hand on a receptacle or at a certain location.

\textbf{Real-robot Setup.}
We use a mobile manipulator that comprises four major components: (i) A four-wheeled differential drive mobile base equipped with an RGB-D camera, a Lidar and an IMU unit; (ii) A 7-DoF Kinova Gen3 robot arm with a Robotic parallel gripper and a RGB-D camera on its wrist.

%% file: sec/4_experiment.tex
\definecolor{mycolor}{RGB}{255,100,0}
\section{Experiment}

This section provides details on experimental task settings and the implementation details of baseline methods. We demonstrate the effectiveness of our system through thorough comparisons with baseline methods and highlight the significance of our design by systematically analyzing execution trials of \robot on these tasks.


\subsection{Experimental Setup}

\paragraph{Baseline} We choose a recent method, \ac{cap}~\cite{liang2023code} as our baseline. \ac{cap} leverages \acs{llm} to generate code to command robots in complex tasks. As the original \ac{cap} method assumes a fully-observable tabletop environment, we augment \ac{cap} with our designed exploration and navigation API functions to build a stronger baseline for comparison, referred as \ac{cap}*.

\begin{figure*}[t]
  \centering
  \includegraphics[width=0.9\linewidth]{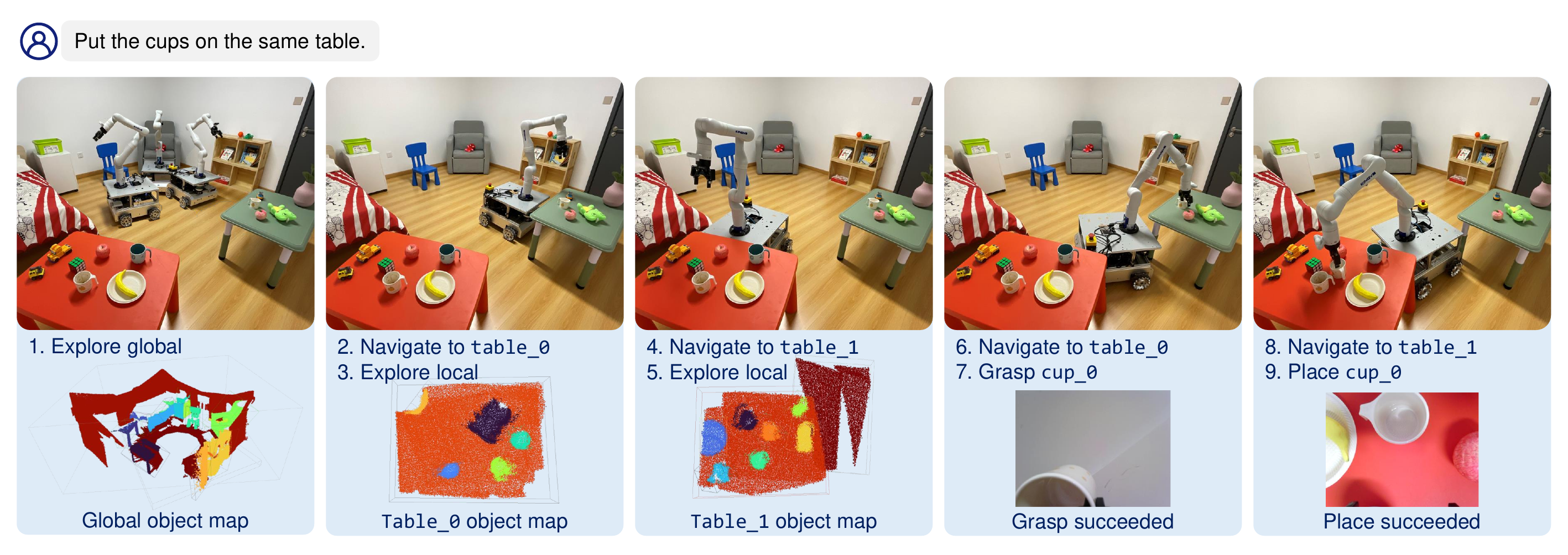}
  \caption{\textbf{A step-by-step visualization of \robot's task execution in \textsc{Gather Cups}.} With the query "Put the cups on the same table." The robot builds a global object map and locates two tables. It then navigates to \texttt{table\_0}, explores locally and identifies one cup on the table. It continues to inspect \texttt{table\_1} and identifies two cups. With situated commonsense reasoning, \robot decides to move the cup from \texttt{table\_0} to \texttt{table\_1} as it is more efficient. It thus navigates back to \texttt{table\_0}, grasps the cup, and verifies the success of grasp with the wrist camera. Finally, it navigates back to \texttt{table\_1} to place the cup down. With the placement once again verified, the task is considered complete.
    }
  \label{fig:step}
\end{figure*}

\begin{table*}[t]
    \begin{minipage}{0.49\linewidth}
        \caption{\textbf{Quantitative results on mobile manipulation.}}
        \centering
        \resizebox{1.\linewidth}{!}{
            \begin{tabular}{cccccccc}
                \toprule
                    \multirow{2}[2]{*}{\textbf{Mobile Task}} & \multicolumn{2}{c}{CaP*} & \multicolumn{3}{c}{\robot}\\ 
                \cmidrule(r){2-3} \cmidrule(r){4-6} 
                    & SR & SSR & SR & SSR & RR \\ 
                \midrule
                    \textsc{Move Toy}
                    &   2 / 5   &  13 / 20  &  \textbf{3 / 5}   &  \textbf{17 / 20} & 2 / 4\\
                    \textsc{Transfer All Toys}       
                    &   1 / 5   &  24 / 42  &  \textbf{2 / 5}   &  \textbf{30 / 42} & 1 / 4\\
                    \textsc{Move Cup and Toy}       
                    &   1 / 5   &  17 / 30  &  \textbf{4 / 5}   &  \textbf{27 / 30} & 4 / 5\\
                    \textsc{Gather Cups}
                    &   2 / 5   &  22 / 33  &  \textbf{4 / 5}   &  \textbf{27 / 30} & 7 / 10 \\
                 \midrule
                    \textbf{Total}
                    &   6 / 20   &  76 / 125  &  \textbf{13/20}   &  \textbf{101 / 122} & 14 / 23\\
                \bottomrule
            \end{tabular}
        }
        \label{tab:mobile}
    \end{minipage}
    \hfill
    \begin{minipage}{0.49\linewidth}
        \caption{\textbf{Quantitative results on tabletop manipulation.}}
        \centering
        \resizebox{1.\linewidth}{!}{
            \begin{tabular}{ccccccc}
                \toprule
                    \multirow{2}[2]{*}{\textbf{Tabletop Task}} & \multicolumn{2}{c}{CaP*} & \multicolumn{3}{c}{\robot}\\ 
                \cmidrule(r){2-3} \cmidrule(r){4-6} 
                    & SR & SSR & SR & SSR & RR \\ 
                
                \midrule
                    \textsc{Place Fruit}       
                    &   5 / 10   &  30 / 40   &  \textbf{7 / 10}   &  \textbf{34 / 40}  & 4 / 7 \\
                    \textsc{Fruit among Cups}      
                    &   6 / 10   &  13 / 20    &  \textbf{8 / 10}   &  \textbf{18 / 20} & 1 / 3 \\
                    \textsc{Prepare Cup}       
                    &   4 / 10   &  12 / 20   &  \textbf{8 / 10}   &  \textbf{17 / 20}  & 7 / 10 \\
                    \textsc{Tidy Table}       
                    &   4 / 10   &  43 / 58    &  \textbf{7 / 10}   &  \textbf{54 / 60} & 5 / 9 \\
                \midrule
                    \textbf{Total}
                    &   19 / 40   &  98 / 138  &  \textbf{30/40}   &  \textbf{123 / 140} & 17 / 29\\
                \bottomrule
            \end{tabular}
        }
        \label{tab:tabletop}
    \end{minipage}
\end{table*}

\begin{table}[t]
\caption{\textbf{Statistics of \robot's failure and recovery.} We use ``DF.'' to denote failures that lead to direct failure of the task, ``R-RR.'' to denote the recovery rate of the re-planning steps.}
\centering
\resizebox{0.9\linewidth}{!}{
    \begin{tabular}{ccccc}
        \toprule
            Failure Type & Reason & DF. & R-RR. & Total \\ 
        \midrule
            \multirow{3}{*}{\textbf{Perception}}
                & false positive 
                    & 3 &  5/5  & 5/8  \\
                & missed detection 
                    & 1 &  1/1  & 1/2  \\
                & visual feedback error
                    & 2 &  1/1  & 1/3  \\
        \midrule
            \multirow{4}{*}{\textbf{Execution}}
                & API call error
                    & 3 &  5/5  & 5/8  \\
                & grasp failed
                    & 1  &  17/23  & 17/24\\
                & place failed
                    & 3  & 3/3     & 3/6 \\
                & navigation failed
                    & 1  & 0/0     & 0/1    \\
                
        \midrule
            Total &
            & 14 &   31/38    &  31/52   \\
        \bottomrule
    \end{tabular}
}

\label{tab:failure case}
\end{table}

\paragraph{Task Design} We meticulously design 4 challenging mobile manipulation tasks with various horizons in a real-world room to evaluate \robot's capability for \ac{ovmm}. The room contains diverse furniture, including several tables, a bed, a sofa, and various objects. 
We design tasks that aim to verify different capabilities of \robot's framework. Specifically, the mobile manipulation tasks are:
\begin{enumerate}[label=A\arabic*,noitemsep,nolistsep]
    \item \textsc{Move Toy}: 
    The task is to move the toy from the table to the bed with clear instructions. This task tests the \textit{basic mobile manipulation capabilities} of robots.
    \item \textsc{Transfer All Toys}: 
    The task is to move scattered toys from different tables to the sofa, which targets robots' \textit{object search capability} in the environment.
    
    \item  \textsc{Move Cup and Toy}:
    The task is to find a cup with specific visual attributes, place it on a plate, and move a toy to the bed for evaluating robots' ability in \textit{sequential tasks and visual reasoning}.

    \item \textsc{Gather Cups}:
    The task is to get all water cups from multiple tables and place them on a single table without specifying which table, for assessing robots' ability to \textit{make efficient plans} when selecting the target table.
    
\end{enumerate}
Furthermore, We designed 4 additional tabletop experiments, closely aligned with the CaP* setup, to more accurately validate the effectiveness of our mechanism. We design the following tasks:
\begin{enumerate}[label=B\arabic*,noitemsep,nolistsep]
    \item \textsc{Place Fruit}: 
    The table is initialized with several fruits, a plate, and other items. The task is to put fruits onto the plate with clear instructions to test the robot's \textit{basic execution and instruction-following} ability.
    
    \item \textsc{Fruit among Cups}:
    The table is initialized with a fruit, several cups, and other items. The task is to place a fruit in the middle of all cups, which tests robots' \textit{understanding of spatial concepts} in instructions. 
    
    \item \textsc{Prepare Cup}: 
    The table is initialized with an empty cup, a used cup, a plate, and other items. The task is to pick up the clean, unused cup and put it onto the plate for pouring water, 
    requiring \textit{commonsense reasoning} for identifying the targeting object.
    
    \item  \textsc{Tidy Table}:
    The table is cluttered with storage boxes and objects (\eg, fruits, toys, \etc). The task is to place these objects into their corresponding storage boxes. 
    We provide instructions like "Can you help me tidy up the table?", to further test robots' \textit{concept understanding and reasoning capability with free-form instructions}.
\end{enumerate}

\begin{figure*}[t]
  \centering
  \includegraphics[width=0.9\linewidth]{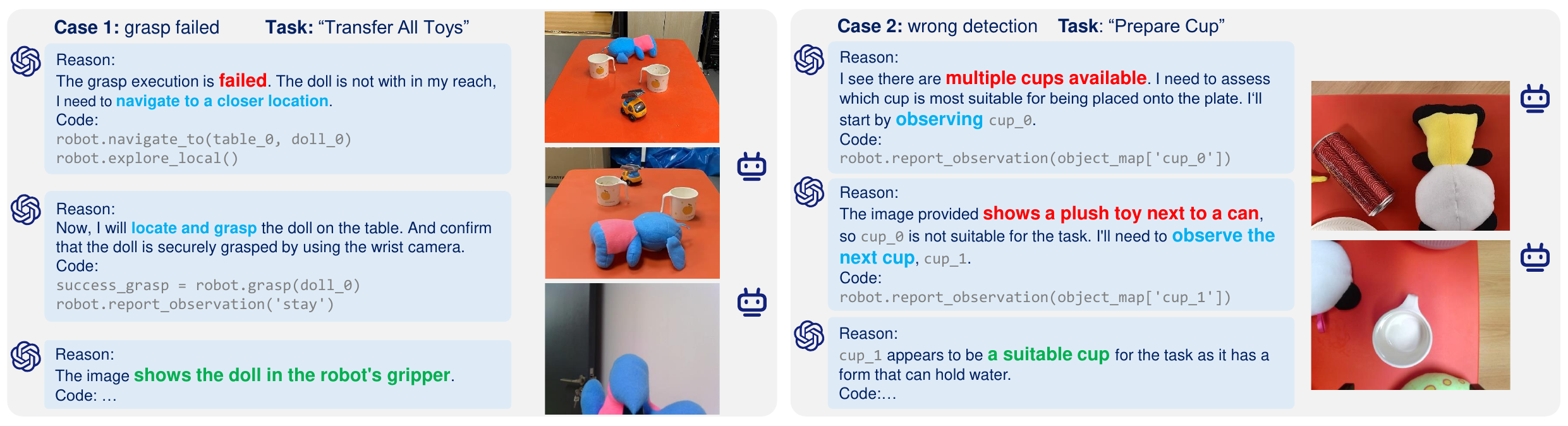}
  \caption{Two examples of recovery from failures: Case 1 demonstrates recovering from a failed grasp attempt by adjusting grasping position. Case 2 describes a scenario of false positive detection, and recover through visual feedback.}
  \label{fig:failure}
\end{figure*}

\paragraph{Experiment Setting} For mobile manipulation tasks, we conduct 5 trials for each task. 
For tabletop tasks, we conduct 10 experiment trials for each task. Similarly, we adjust the scene configuration between trials by introducing variations of objects' types, their arrangements, and quantities on the table.

\paragraph{Metrics} We report the success rate (SR) of goal completion and the step-wise success rate (SSR) of action execution for all models. Each execution API call is considered one step (\ie, excluding perception APIs like \texttt{report\_observation}), and we count the number of successful ones over all calls for SSR. Different planned paths and methods may require varying numbers of steps for the same task, especially considering \robot's replanning mechanism. Additionally, to unveil \robot's ability in recovering from failure, we report the recovery rate (RR) of \robot by tallying all replanned executions and the successful ones within these executions.

\subsection{Experimental Results and Analyses}\label{sec:exp:results}

As shown in~\cref{tab:mobile} and~\cref{tab:tabletop}, \robot achieves consistent and significant improvements in goal completion for both the mobile manipulation and the tabletop setting. Specifically, \robot achieves an overall success rate of 65\% (13/20) on the mobile manipulation setting, outperforming the \ac{cap}* baseline (30\%, 6/20) by 35\%. Similarly, under the tabletop setting, \robot achieves a success rate of 75\%, significantly outperforming baseline (47.5\%, 19/40) by 27.5\%.
Additionally, the ability of \robot to recover from failures significantly enhances its stepwise success rate by identifying the failure step and replanning, demonstrating a higher performance of $101/122$ compared to CaP*'s $76/125$ and $123/140$ compared to $98/138$. This improvement also contributes to the higher overall success rate. These quantitative results validate that \robot, equipped with the ability to replan using closed-loop feedback, effectively identifies and corrects errors encountered during task execution. This capability facilitates task execution and goal achievements in challenging real-world tasks.

\subsection{Failure Analysis}
As shown in~\cref{tab:mobile} and~\cref{tab:tabletop}, we demonstrate the effectiveness of failure recovery mechanism from both  
recovering steps and recovery rate. 
In this section, we provide analysis to systematically categorize \robot's failure cases and highlight how \robot recovers from such failures utilizing closed loop feedback.

\paragraph{Perceptual Failures} Perception failures are primarily caused by detection errors during exploration. \robot can use object-level perception for close inspection and utilize visual feedback for solving the missed or wrong detection problem. For missed detection, \robot directs perception modules to rebuild the local object scene graph and re-detect the missing object, achieving a 100\% recover rate as shown in~\cref{tab:failure case}. For wrong detection, \robot utilizes GPT-4V to conduct a verification step for detected objects. For example, 
when local-level perception module detects multiple candidate cups, \robot verifies each cup with image observations and finds that \texttt{cup\_0} is actually a doll that is wrongly detected as cup and not related to the task. Though this verification process can help mitigate the problem, it is still error-prone to incorrect predictions, leaving 6 falsely detected objects after verification, with three of which lead to task failure as shown in~\cref{tab:failure case}. 
\paragraph{Execution Failures} \robot's GPT-4V-based planning method may sometimes generate incorrect plans or invalid API calls, such as attempting to place an object without prior grasping or calling the navigation function with an object name instead of an object. For these errors, \robot verifies the generated plan and code, and triggers exceptions during execution, providing explicit feedback indicating the missing step or wrong function call for GPT-4V to rectify the plan. For actual execution, the primary source of failure is caused by unsuccessful grasps. Grasping failures are primarily caused by impractical position the robot navigates to which significantly constrains its space for manipulation (\eg corner of the table, or close to the wall). The case 1 of \cref{fig:failure} shows an example of recovering from grasp failure. 

    
    


\subsection{Discussions}

\textbf{Is commonsense reasoning using LLMs enough for mobile manipulation with open-ended instructions?}
We argue that completing these tasks requires both the commonsense knowledge provided by LLMs and the ability to interactively explore and update scene information. 
Taking the \textsc{Prepare Cup} (B3)  task as an example, 
when multiple cups are detected, the robot must compare them to identify the most suitable one for the task. Crafting a plan solely based on the instruction becomes challenging without understanding the status of all cups. 

\textbf{Why is replanning with closed-loop feedback important for robot manipulation tasks?}
Compared to tabletop manipulation tasks, mobile manipulation tasks involve longer sequences, requiring the robot to first explore the room and then shuttle between furniture to complete cross-furniture manipulation tasks. As shown in~\cref{tab:mobile}, mobile manipulation tasks require a significantly higher average execution step. Meanwhile, the long execution sequence unveils the effectiveness of replanning in \robot compared to \ac{cap}* as long sequences increase the likelihood of execution failure. 
Replanning with closed-loop feedback enables \robot to detect failures and attempt to recover from them, thereby 
reducing the likelihood of task failure. 

%% file: sec/5_conclusion.tex
\section{Conclusion}

In conclusion, we present \robot, a novel closed-loop framework integrating GPT-4V with robust robotic primitives for open-vocabulary mobile manipulation. Real-world experiments show its superior ability to interpret open-ended instructions, reason over multi-modal feedback, and recover from perception and execution failures. Leveraging GPT-4V's reasoning capabilities, \robot achieves unprecedented flexibility and intelligence in \ac{ovmm} tasks. 
We hope this work inspires further research on integrating foundation models with robotics to enhance intelligence and autonomy.

%% file: root.bbl
\begin{thebibliography}{10}
\providecommand{\url}[1]{#1}
\csname url@rmstyle\endcsname
\providecommand{\newblock}{\relax}
\providecommand{\bibinfo}[2]{#2}
\providecommand\BIBentrySTDinterwordspacing{\spaceskip=0pt\relax}
\providecommand\BIBentryALTinterwordstretchfactor{4}
\providecommand\BIBentryALTinterwordspacing{\spaceskip=\fontdimen2\font plus
\BIBentryALTinterwordstretchfactor\fontdimen3\font minus \fontdimen4\font\relax}
\providecommand\BIBforeignlanguage[2]{{%
\expandafter\ifx\csname l@#1\endcsname\relax
\typeout{** WARNING: IEEEtran.bst: No hyphenation pattern has been}%
\typeout{** loaded for the language `#1'. Using the pattern for}%
\typeout{** the default language instead.}%
\else
\language=\csname l@#1\endcsname
\fi
#2}}

\bibitem{bommasani2021opportunities}
R.~Bommasani, D.~A. Hudson, E.~Adeli, R.~Altman, S.~Arora, S.~von Arx, M.~S. Bernstein, J.~Bohg, A.~Bosselut, E.~Brunskill, \emph{et~al.}, ``On the opportunities and risks of foundation models,'' \emph{arXiv preprint arXiv:2108.07258}, 2021.

\bibitem{brohan2023rt}
A.~Brohan, N.~Brown, J.~Carbajal, Y.~Chebotar, X.~Chen, K.~Choromanski, T.~Ding, D.~Driess, A.~Dubey, C.~Finn, \emph{et~al.}, ``Rt-2: Vision-language-action models transfer web knowledge to robotic control,'' \emph{arXiv preprint arXiv:2307.15818}, 2023.

\bibitem{brohan2022rt}
A.~Brohan, N.~Brown, J.~Carbajal, Y.~Chebotar, J.~Dabis, C.~Finn, K.~Gopalakrishnan, K.~Hausman, A.~Herzog, J.~Hsu, \emph{et~al.}, ``Rt-1: Robotics transformer for real-world control at scale,'' \emph{arXiv preprint arXiv:2212.06817}, 2022.

\bibitem{brown2020language}
T.~Brown, B.~Mann, N.~Ryder, M.~Subbiah, J.~D. Kaplan, P.~Dhariwal, A.~Neelakantan, P.~Shyam, G.~Sastry, A.~Askell, \emph{et~al.}, ``Language models are few-shot learners,'' \emph{Advances in Neural Information Processing Systems (NeurIPS)}, vol.~33, pp. 1877--1901, 2020.

\bibitem{curtis2022long}
A.~Curtis, X.~Fang, L.~P. Kaelbling, T.~Lozano-P{\'e}rez, and C.~R. Garrett, ``Long-horizon manipulation of unknown objects via task and motion planning with estimated affordances,'' in \emph{International Conference on Robotics and Automation (ICRA)}.\hskip 1em plus 0.5em minus 0.4em\relax IEEE, 2022, pp. 1940--1946.

\bibitem{ding2023integrating}
Y.~Ding, X.~Zhang, S.~Amiri, N.~Cao, H.~Yang, A.~Kaminski, C.~Esselink, and S.~Zhang, ``Integrating action knowledge and llms for task planning and situation handling in open worlds,'' \emph{Autonomous Robots}, 2023.

\bibitem{du2022survey}
Y.~Du, Z.~Liu, J.~Li, and W.~X. Zhao, ``A survey of vision-language pre-trained models,'' \emph{arXiv preprint arXiv:2202.10936}, 2022.

\bibitem{garrett2020online}
C.~R. Garrett, C.~Paxton, T.~Lozano-P{\'e}rez, L.~P. Kaelbling, and D.~Fox, ``Online replanning in belief space for partially observable task and motion problems,'' in \emph{International Conference on Robotics and Automation (ICRA)}.\hskip 1em plus 0.5em minus 0.4em\relax IEEE, 2020, pp. 5678--5684.

\bibitem{guo2023doremi}
Y.~Guo, Y.-J. Wang, L.~Zha, Z.~Jiang, and J.~Chen, ``Doremi: Grounding language model by detecting and recovering from plan-execution misalignment,'' \emph{arXiv preprint arXiv:2307.00329}, 2023.

\bibitem{han2024interpret}
M.~Han, Y.~Zhu, S.-C. Zhu, Y.~N. Wu, and Y.~Zhu, ``Interpret: Interactive predicate learning from language feedback for generalizable task planning,'' \emph{arXiv preprint arXiv:2405.19758}, 2024.

\bibitem{hu2023look}
Y.~Hu, F.~Lin, T.~Zhang, L.~Yi, and Y.~Gao, ``Look before you leap: Unveiling the power of gpt-4v in robotic vision-language planning,'' \emph{arXiv preprint arXiv:2311.17842}, 2023.

\bibitem{huang2022inner}
W.~Huang, F.~Xia, T.~Xiao, H.~Chan, J.~Liang, P.~Florence, A.~Zeng, J.~Tompson, I.~Mordatch, Y.~Chebotar, \emph{et~al.}, ``Inner monologue: Embodied reasoning through planning with language models,'' \emph{arXiv preprint arXiv:2207.05608}, 2022.

\bibitem{jiao2021efficieint}
Z.~Jiao, Z.~Zeyu, W.~Wang, D.~Han, S.-C. Zhu, Y.~Zhu, and H.~Liu, ``Efficient task planning for mobile manipulation: a virtual kinematic chain perspective,'' in \emph{International Conference on Intelligent Robots and Systems (IROS)}, 2021.

\bibitem{kim2024openvla}
M.~J. Kim, K.~Pertsch, S.~Karamcheti, T.~Xiao, A.~Balakrishna, S.~Nair, R.~Rafailov, E.~Foster, G.~Lam, P.~Sanketi, \emph{et~al.}, ``Openvla: An open-source vision-language-action model,'' \emph{arXiv preprint arXiv:2406.09246}, 2024.

\bibitem{liang2023code}
J.~Liang, W.~Huang, F.~Xia, P.~Xu, K.~Hausman, B.~Ichter, P.~Florence, and A.~Zeng, ``Code as policies: Language model programs for embodied control,'' in \emph{International Conference on Robotics and Automation (ICRA)}, 2023.

\bibitem{liu2024ok}
P.~Liu, Y.~Orru, C.~Paxton, N.~M.~M. Shafiullah, and L.~Pinto, ``Ok-robot: What really matters in integrating open-knowledge models for robotics,'' \emph{arXiv preprint arXiv:2401.12202}, 2024.

\bibitem{liu2023reflect}
Z.~Liu, A.~Bahety, and S.~Song, ``Reflect: Summarizing robot experiences for failure explanation and correction,'' in \emph{Conference on Robot Learning (CoRL)}.\hskip 1em plus 0.5em minus 0.4em\relax PMLR, 2023, pp. 3468--3484.

\bibitem{mei2024replanvlm}
A.~Mei, G.-N. Zhu, H.~Zhang, and Z.~Gan, ``Replanvlm: Replanning robotic tasks with visual language models,'' \emph{arXiv preprint arXiv:2407.21762}, 2024.

\bibitem{gpt4v}
OpenAI, ``Gpt-4v(ision) system card,'' \url{https://cdn.openai.com/papers/GPTV_System_Card.pdf}, 2023.

\bibitem{qiu2024open}
D.~Qiu, W.~Ma, Z.~Pan, H.~Xiong, and J.~Liang, ``Open-vocabulary mobile manipulation in unseen dynamic environments with 3d semantic maps,'' \emph{arXiv preprint arXiv:2406.18115}, 2024.

\bibitem{radford2021clip}
A.~Radford, J.~W. Kim, C.~Hallacy, A.~Ramesh, G.~Goh, S.~Agarwal, G.~Sastry, A.~Askell, P.~Mishkin, J.~Clark, \emph{et~al.}, ``Learning transferable visual models from natural language supervision,'' in \emph{International conference on machine learning}.\hskip 1em plus 0.5em minus 0.4em\relax PMLR, 2021, pp. 8748--8763.

\bibitem{rana2023sayplan}
K.~Rana, J.~Haviland, S.~Garg, J.~Abou-Chakra, I.~Reid, and N.~Suenderhauf, ``Sayplan: Grounding large language models using 3d scene graphs for scalable task planning,'' \emph{arXiv preprint arXiv:2307.06135}, 2023.

\bibitem{ravi2024sam}
N.~Ravi, V.~Gabeur, Y.-T. Hu, R.~Hu, C.~Ryali, T.~Ma, H.~Khedr, R.~R{\"a}dle, C.~Rolland, L.~Gustafson, \emph{et~al.}, ``Sam 2: Segment anything in images and videos,'' \emph{arXiv preprint arXiv:2408.00714}, 2024.

\bibitem{singh2023progprompt}
I.~Singh, V.~Blukis, A.~Mousavian, A.~Goyal, D.~Xu, J.~Tremblay, D.~Fox, J.~Thomason, and A.~Garg, ``Progprompt: Generating situated robot task plans using large language models,'' in \emph{International Conference on Robotics and Automation (ICRA)}.\hskip 1em plus 0.5em minus 0.4em\relax IEEE, 2023, pp. 11\,523--11\,530.

\bibitem{skreta2024replan}
M.~Skreta, Z.~Zhou, J.~L. Yuan, K.~Darvish, A.~Aspuru-Guzik, and A.~Garg, ``Replan: Robotic replanning with perception and language models,'' \emph{arXiv preprint arXiv:2401.04157}, 2024.

\bibitem{wake2023gpt}
N.~Wake, A.~Kanehira, K.~Sasabuchi, J.~Takamatsu, and K.~Ikeuchi, ``Gpt-4v (ision) for robotics: Multimodal task planning from human demonstration,'' \emph{arXiv preprint arXiv:2311.12015}, 2023.

\bibitem{wang2023voyager}
G.~Wang, Y.~Xie, Y.~Jiang, A.~Mandlekar, C.~Xiao, Y.~Zhu, L.~Fan, and A.~Anandkumar, ``Voyager: An open-ended embodied agent with large language models,'' \emph{arXiv preprint arXiv:2305.16291}, 2023.

\bibitem{wang2023describe}
Z.~Wang, S.~Cai, G.~Chen, A.~Liu, X.~Ma, and Y.~Liang, ``Describe, explain, plan and select: interactive planning with llms enables open-world multi-task agents,'' in \emph{Advances in Neural Information Processing Systems (NeurIPS)}, 2023.

\bibitem{wolfe2010combined}
J.~Wolfe, B.~Marthi, and S.~Russell, ``Combined task and motion planning for mobile manipulation,'' in \emph{Proceedings of the International Conference on Automated Planning and Scheduling}, vol.~20, 2010, pp. 254--257.

\bibitem{yao2022react}
S.~Yao, J.~Zhao, D.~Yu, N.~Du, I.~Shafran, K.~R. Narasimhan, and Y.~Cao, ``React: Synergizing reasoning and acting in language models,'' in \emph{International Conference on Learning Representations (ICLR)}, 2022.

\bibitem{yenamandra2023homerobot}
S.~Yenamandra, A.~Ramachandran, K.~Yadav, A.~Wang, M.~Khanna, T.~Gervet, T.-Y. Yang, V.~Jain, A.~W. Clegg, J.~Turner, \emph{et~al.}, ``Homerobot: Open-vocabulary mobile manipulation,'' \emph{arXiv preprint arXiv:2306.11565}, 2023.

\bibitem{yokoyama2023asc}
N.~Yokoyama, A.~Clegg, J.~Truong, E.~Undersander, T.-Y. Yang, S.~Arnaud, S.~Ha, D.~Batra, and A.~Rai, ``Asc: Adaptive skill coordination for robotic mobile manipulation,'' \emph{IEEE Robotics and Automation Letters (RA-L)}, vol.~9, no.~1, pp. 779--786, 2023.

\bibitem{zeng2023large}
F.~Zeng, W.~Gan, Y.~Wang, N.~Liu, and P.~S. Yu, ``Large language models for robotics: A survey,'' \emph{arXiv preprint arXiv:2311.07226}, 2023.

\end{thebibliography}
